\documentclass[twoside,leqno,twocolumn]{article}

\usepackage[letterpaper]{geometry}

\usepackage{ltexpprt}
\usepackage{hyperref}

\usepackage{amsmath, amssymb, latexsym}
\usepackage[dvips]{epsfig,psfrag}
\usepackage{url}
\usepackage{color}
\definecolor{darkred}{RGB}{150,0,0}
\newcommand{\ssg}[1]{\textcolor{black}{#1}}

\begin{document}

\newcommand\relatedversion{}

\title{\Large Profiling checkpointing schedules in adjoint ST-AD}
\author{Laurent Hasco\"{e}t \thanks{INRIA, Sophia-Antipolis, France}
\and Jean-Luc Bouchot \footnotemark[1]
\and Shreyas Sunil Gaikwad \thanks{Oden Institute for Computational Engineering and Sciences, The University of Texas at Austin, USA}
\and Sri Hari Krishna Narayanan \thanks{Argonne National Laboratory, Lemont, IL, USA}
\and Jan H\"{u}ckelheim \footnotemark[3]}

\date{}

\maketitle

% Copyright Statement
% When submitting your final paper to a SIAM proceedings, it is requested that you include
% the appropriate copyright in the footer of the paper.  The copyright added should be
% consistent with the copyright selected on the copyright form submitted with the paper.
% Please note that "20XX" should be changed to the year of the meeting.

% Default Copyright Statement
\fancyfoot[R]{\scriptsize{Copyright \textcopyright\ 20XX by SIAM\\
Unauthorized reproduction of this article is prohibited}}

% Depending on which copyright you agree to when you sign the copyright form, the copyright
% can be changed to one of the following after commenting out the default copyright statement
% above.

%\fancyfoot[R]{\scriptsize{Copyright \textcopyright\ 20XX\\
%Copyright for this paper is retained by authors}}

%\fancyfoot[R]{\scriptsize{Copyright \textcopyright\ 20XX\\
%Copyright retained by principal author's organization}}

%\pagenumbering{arabic}
%\setcounter{page}{1}%Leave this line commented out.

\begin{abstract} \small\baselineskip=9pt
Checkpointing is a cornerstone of data-flow reversal in adjoint algorithmic differentiation. Checkpointing is a storage/recomputation trade-off that can be applied at different levels, one of which being the call tree. We are looking for good placements of checkpoints onto the call tree of a given application, to reduce run time and memory footprint of its adjoint. There is no known optimal solution to this problem other than a combinatorial search on all placements. We propose a heuristics based on run-time profiling of the adjoint code. We describe implementation of this profiling tool in an existing source-transformation AD tool. We demonstrate the interest of this approach on test cases taken from the MITgcm ocean and atmospheric global circulation model. We discuss the limitations of our approach and propose directions to lift them.\end{abstract}

\section{Introduction}\label{secIntro}

Source-transformation algorithmic differentiation (ST-AD) in its adjoint mode transforms a {\it primal} code that evaluates some original function into an {\it adjoint} code that computes its gradient.
It is well known~\cite{Griewank2008EDP} that the most efficient implementation of the adjoint code must progress backwards of the original computation, progressively using values originating from the primal execution.
The amount of values used grows linearly with the run time of the primal code and, since they are used in the reverse of their production order, their management ({\it data-flow reversal}) is a key issue that requires a delicate trade-off between storage and recomputation.

This work focuses on one particular setting, where data-flow reversal is primarily done through a stack and the memory cost of this stack is mitigated through a classical storage/recomputation trade-off known as {\it checkpointing}.
Tuning this trade-off is difficult, and optimal tuning exists only in specific cases~\cite{Griewank2000ARA}.
In our setting, checkpointing can be applied at every procedure call.
When the granularity of procedure calls is not fine enough, checkpointing can also be applied at user-designated procedure fragments, provided they {\em could} be turned into procedures.
Potential checkpointing locations therefore end up being the nodes of the call graph, possibly extended to procedure fragments, and at run time they are nested as the nodes of the call tree.
In this work, we examine the question of finding a subset of the checkpointing locations that will result in good enough execution time given a limited storage budget.
Optimality, although desirable, has already been shown to be NP-hard~\cite{NaumannDAGNP09}.
We will rather explore heuristics, based on profiling results, to decide which checkpoint locations should be activated or inhibited to achieve a better performance.

In section~\ref{secCkpModel}, we describe in more detail the principle of checkpointing in the setting of our stack-based data-flow reversal, and illustrate its costs and benefits.
Section~\ref{secOtherCkp} contrasts with checkpointing in other settings and relates this work with other profiling support in existing AD tools.
Section~\ref{secProfiling} describes the information appropriate to guide the choice of activated checkpoints, and how to gather it from a profiled run.
Section~\ref{secImplementation} discusses implementation of this profiling in an existing source-transformation AD tool, and section~\ref{secApplication} applies it to two realistic test-cases taken from the MITgcm code suite. We will show how the developer can achieve a significant performance gain by exploiting the profiling results.
In section~\ref{secFurther}, we come back to some limitations of our proposed approach and discuss how they could be overcome, before concluding in section~\ref{secConcl}.

%The contributions of this paper are:
%\begin{itemize}
%\item
%\end{itemize}

%This work has the following limitations:
%\begin{itemize}
%\item
%\end{itemize}

\section{Our checkpointing model / setting}\label{secCkpModel}

In our setting, data-flow reversal is achieved by storing intermediate values of the primal execution. Consequently, the adjoint code basically obeys a two-sweeps structure:
\begin{enumerate}
    \item a first, ``{\em forward}'', sweep runs the primal code, augmented to organize the storage of whatever intermediate values or derivatives will be needed in the second sweep.
    \item a second, ``{\em backward}'', sweep propagates the gradients in reverse order, retrieving the intermediate values from the first sweep when needed.
\end{enumerate}
% {\it \color{darkred} [SSG]: I feel it would look better if only half a bullet point was not on a new page. [LLH] right. Anyone knows how to do this?}

Naturally, the appropriate storage structure for that is a stack. Note that the intermediate values that are not used in the derivative computation, such as those appearing only in linear computations, need not be stored. This can reduce the stack size significantly.
AD tools implement several kinds of data-flow analysis to detect many opportunities to reduce the stack usage. Still, stack size generally grows linearly with run-time and the checkpointing technique, which trades recomputation for peak stack size, is unavoidable.

\begin{figure}
\includegraphics[width=\linewidth]{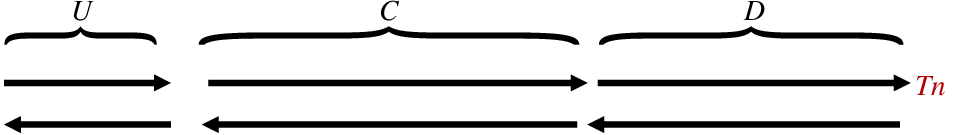}
\caption{Sketch of execution for the adjoint of a code arbitrarily split as a sequence of three parts $U$, $C$, and $D$. Thick arrows to the right stand for forward sweeps, running the primal code together with storing intermediate values on the stack. Thick arrows to the left stand for backward sweeps, that retrieve values from the stack and use them while propagating gradients backwards. Time goes top-down, and inside each line time follows the arrows direction.}
 \label{figBasisModel}
\end{figure}

Figure~\ref{figBasisModel} sketches execution of the adjoint code, in the base case where no checkpointing is active. Code fragment $C$, a candidate for future checkpointing, is singled out from upstream and downstream code $U$ and $D$, only for future comparison with the case where checkpointing is applied to $C$. At the turn point between the forward and backward sweeps, the stack reaches its maximum size $\hbox{\it Tn}$.

\begin{figure}
\includegraphics[width=\linewidth]{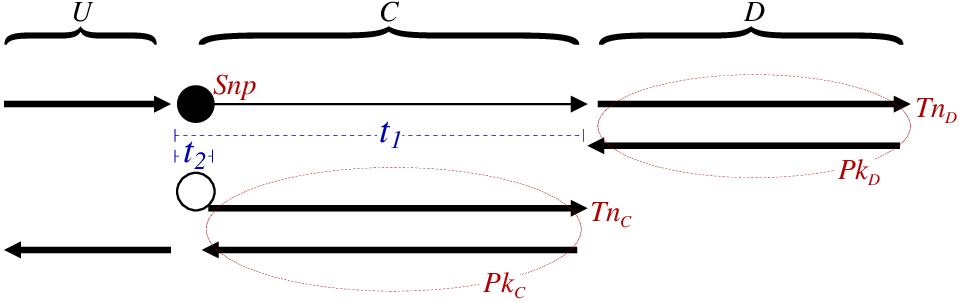}
\caption{Sketch of execution for the adjoint of the same code than in Fig.~\ref{figBasisModel}, this time checkpointing code part $C$. Time goes top-down, and inside each line time follows the arrows direction. The thin arrow to the right stands for the extra execution of the primal $C$, and the black and white bullets stand for respectively the writing and reading of the {\em snapshot}, i.e. a set of variables sufficient for identical duplicate execution of $C$.}
\label{figCheckpointModel}
\end{figure}

Checkpointing reduces the peak stack size by allowing for one recomputation of a chosen fragment of the primal code, here $C$ as shown in Fig.~\ref{figCheckpointModel}.
At the cost of this duplicate execution of primal fragment $C$, which does not write values to the stack, the stack size reaches two local peaks $\hbox{\it Tn}_D$ and $\hbox{\it Tn}_C$, both hopefully smaller than $\hbox{\it Tn}$.
On the one hand, $\hbox{\it Tn}_D$ contains intermediate values from $U$, $D$, but not from $C$, plus a so-called ``{\em snapshot}'' needed to run $C$ again and which we hope is smaller than the intermediate values in $C$.
On the other hand, $\hbox{\it Tn}_C$ only contains intermediate values from $U$ and $C$. Notice that $\hbox{\it Tn}_C$ does not contain the snapshot, which has been read to prepare for the forward sweep of $C$ and immediately deleted. Symbols $\hbox{\it Pk}_D$, $\hbox{\it Pk}_C$, and $Snp$ in Fig.~\ref{figCheckpointModel} denote stack sizes that will be discussed in section~\ref{secProfiling}.
It is worth observing on Fig.~\ref{figCheckpointModel} that the reading and writing of the snapshot blend nicely with the general stack behavior, which allows us to store the snapshot and the other intermediate values onto the same stack. Whatever efficient implementation, or cache level optimization, exists for the stack will be equally beneficial for snapshots.
Also notice that \textit{some} values in the snapshot may have been already stored in $U$, not too deep in the stack. Although some data-flow analysis can detect this and therefore slightly reduce either the snapshot or the stack after $U$, our experiments have shown that the benefit is very often minor. As discussed in section~\ref{secFurther}, this sort of extra improvement, implemented in the AD tool we are using here, can be the cause for inaccuracy in the profiling predictions.

Notice that on the figures, the position and length of the arrows do not represent the stack size, but only their correspondence with the primal code. In other words $\hbox{\it Tn}_C$ may well be larger than $\hbox{\it Tn}_D$.

On large codes, checkpointing can and must be applied repeatedly, recursively inside fragments $U$, $C$, and $D$.
With a well-balanced choice of the nested checkpoints, i.e. the choice of $C$ and of all other checkpoints nested inside $U$, $C$, and $D$, one can achieve logarithmic growth -- with respect to the primal run time -- of both the peak stack size and of the number of times a given primal code fragment is repeatedly run.
Notice however that the control structure of the primal code restricts the choice of the nested checkpoints. Basically since $C$ will be re-executed, it must have unique entry and exit points.
It must also be ``{\em reentrant}'', i.e. contains no adverse side-effect such as an isolated {\tt send} or {\tt receive}, {\tt malloc} or {\tt free}, file I-O, etc.
Obvious checkpointing candidates $C$'s are all the (reentrant) procedure calls, but one may also define at will additional checkpoints inside procedure bodies, essentially when these additional checkpoints {\em could} be turned into reentrant procedure calls. In this work, we will assume that one has defined a large enough collection of nested potential checkpoints, and our goal is to use profiling to selectively inhibit or activate these checkpoints to reach good performance in run time and in peak stack size.

At one extreme, activating all the potential checkpoints (i.e. actually performing checkpointing on them) will severely affect run time due to primal code recomputations.
Conversely, inhibiting all potential checkpoints can easily lead to excessive stack memory use.
One can look for an optimal trade off for instance: for a given fixed stack memory budget, look for the checkpoints activation that minimizes primal code recomputation time.
Previous works~\cite{NaumannDAGNP09} have shown that a general answer to this combinatorial problem is NP-hard, possibly resorting to Integer Programming approaches~\cite{LNM16}. We will here explore a greedy approach, progressively improving our subset of activated checkpoints. Still, the information returned by profiling could also be fed to optimal search tools.

The combined questions we consider are: what information can we extract from run-time profiling of the adjoint code, and how can we use it to guide selection of potential checkpoints towards a good enough trade off?

\section{Related approaches} \label{secOtherCkp}

In the so-called ``{\em recompute-all}'' model, primal values are made available by repeatedly running the primal code from a chosen initial point.
This approach trades computation time for memory space.
The alternative ``{\em store-all}'' model stores the needed primal values on a stack, sparing the extra recomputation time at the cost of storage space.
On realistic codes that run for long times, both models require a similar notion of checkpointing, and the question of the optimal choice of checkpoints is the same in both models. The tool TAF~\cite{Giering1998} (``recompute-all'' model) allows one to fine-tune the checkpointing locations and the variables they must store, through user directives. This manual tuning is (by principle) more powerful than any automated approach, but is labor-intensive. It could as well benefit from some profiling.

In some ``store-all'' models, the values stored on the stack are the intermediate values of variables computed by the primal code, or at least the subset of those that are indeed used in the partial derivatives accumulated during gradient computation.
In other models~\cite{RefOpenAD} they can be the partial derivatives directly.
When storing intermediate values, one can choose to store them upon each primal instruction whose adjoint uses the value. This ``{\em store-on-use}'' has the risk of storing the same value several times. The alternative ``{\em store-on-kill}'', which we follow in the sequel, is to store the value only when it is going to be overwritten by the next primal instruction.

The present work considers checkpointing essentially at the level of the call tree. In the restricted setting of a time-stepping loop, several checkpointing schedules have been studied. In particular the well-known {\em binomial checkpointing}~\cite{Griewank2000ARA} is optimal, in its application case. These checkpointing strategies on loops differ from ours as they amount to choosing the extent of the nested checkpoints over the sequences of time steps, whereas our setting operates on a collection of fixed potential checkpoints on which we search a good subset. Given the nested time-step sequences that binomial checkpointing has found optimal, inhibiting checkpointing on any such sequence can only lead to a less efficient scheme. Conversely we do not consider in our setting the option to enlarge or shrink the code fragments $C$ to which checkpointing may be applied. Binomial checkpointing, or its refinements dedicated to very large snapshots such as H-revolve~\cite{10.1145/3378672}, are independent from the present checkpointing on call trees. We will see in section~\ref{secWithBinomial} how they may complement each other.

%{\it \color{darkred} SHK: H-revolve chooses where to places checkpoints and is a refinement of binomial checkpointing where all forward steps take $u_f$ time and all reverse steps take $u_b$ time. So I am  not sure it belongs in this discussion.}
%H-revolve is a general optimal strategy to execute an Adjoint Computation graph on a hierarchical storage system with increasing reading and writing times for deeper levels of storage~\cite{10.1145/3378672}.

Profiling is one answer to improve performance when no static analysis can guarantee optimality at compile time. Machine learning frameworks such as PyTorch, TensorFlow, and JAX offer automatic differentiation capability as well as profiling. In JAX, a common use of profiling is to understand the GPU or TPU memory usage of a program, for example if trying to debug an out-of-memory problem~\cite{jaxdocs}. A profiler trace can be created by instrumenting the code with the {\tt jax.profiler.start\_trace()} and {\tt jax.profiler.stop\_trace()} methods. The trace can be viewed later using a graphical interface. As opposed to the approach presented here, the profiling information is not used to select a differentiation strategy.

\section{A profiling model for checkpointing}\label{secProfiling}

We will distinguish static and dynamic checkpoint locations. We call static a location in the source code, and more precisely in the primal source code. This is where the end-user may conveniently decide to (de-)activate a checkpoint e.g. through a differentiation directive. Dynamic checkpoints are the run-time occurrences when the execution pointer reaches a static checkpoint. There can be several dynamic checkpoints for each static one. It is in principle possible to apply different checkpointing choices to each dynamic occurrence, but this is impractical and leads to a more involved implementation.  We therefore choose to attach the profiling information that we seek to each static checkpoint, although it will be collected and accumulated at run-time on each of its dynamic occurrences. 

As a starting configuration that we will profile and progressively improve upon, we will activate all potential checkpoints. One reason is that this configuration gives profiling an easier access to useful information. The other reason is that it is less likely to fail at run-time by lack of memory, although it may take longer. It is therefore a safer starting point when the structure of the primal code is not well known. As the improvement process goes on, the checkpointing configuration is modified according to profiling results, and a new profiling run is in theory needed for each configuration in order to determine the next configuration.

For some current configuration, for every potential checkpoint $X$ which is currently activated, we expect profiling to give us the following key figures about inhibiting $X$:
\begin{itemize}
\item the associated run-time benefit $\Delta t(X)$, i.e. the run-time difference when $X$ is switched from activated to inhibited. It is always negative (run-time decreases).
\item the associated memory cost $\Delta \hbox{\it Pk}(X)$, i.e. the peak stack size difference when $X$ is switched from activated to inhibited. It is in general positive except for some extreme cases.
\end{itemize}
Let us look back at Fig.~\ref{figCheckpointModel}, now assuming we are in the general case where there can be nested checkpoints in the adjoint codes for $C$ and $D$ (and $U$). Just view $C$ and $D$ as sections of the primal code, in which there may be any number of checkpointed sections $X$, one of them coinciding with $C$.
Let us call {\it round trip} the sequence of a forward sweep and its associated backward sweep.
For instance the rightmost part of Fig.~\ref{figCheckpointModel} is the round trip on $D$, also written $\overline{D}$, the sequence of its forward sweep $\overrightarrow{D}$ and backward sweep $\overleftarrow{D}$.
The bottom two arrows of the central part of Fig.~\ref{figCheckpointModel} form the round trip on $C$.
A round trip can include nested checkpoints and for instance the round trip on $U;C;D$ is exactly Fig.~\ref{figBasisModel} (if $C$ is inhibited) or Fig.~\ref{figCheckpointModel} (if $C$ is activated).

Since we now assume that there may be nested checkpoints inside $D$, we must distinguish peak stack sizes $\hbox{\it Tn}_D$ and $\hbox{\it Pk}_D$: as defined already, $\hbox{\it Tn}_D$ is the (local) peak reached at the turn point at the end of $\overrightarrow{D}$, whereas $\hbox{\it Pk}_D$ is the (global) maximum peak stack size attained during the round trip on $D$, i.e. at the turn point after $\overrightarrow{D}$ but also at the turn point of any checkpoint $N$ nested in $D$. Nothing prevents $\overrightarrow{N}$ from storing more intermediate values than $D$ and therefore $\hbox{\it Pk}_D \ge \hbox{\it Tn}_D$ in general.
In the sequel we will need to keep track of both $\hbox{\it Tn}_D$ and $\hbox{\it Pk}_D$. The same remark applies to $C$.

%We will accumulate cost/benefit figures bottom-up and right-to-left on the tree of nested dynamic checkpoints or in other words bottom-up on the tree of nested round trips. 
Profiling is performed on the adjoint code corresponding to the current checkpointing configuration. We want profiling to return, for each static checkpoint $X$ that is currently active, the cost/benefits $\Delta t(X)$ and $\Delta \hbox{\it Pk}(X)$. Only as an intermediate information, profiling will also return $\Delta \hbox{\it Tn}(X)$.
As cost/benefits can be evaluated independently for each $X$, we will in the sequel drop the $(X)$ to keep notation light.

We will accumulate cost/benefit figures bottom-up and right-to-left on the tree of nested dynamic checkpoints or in other words, bottom-up on the tree of nested round trips. Following Fig.~\ref{figCheckpointModel}, the key recursive operation is therefore to compute these figures for the round trip on the code sequence $C;D$, given these figures for the round trips on $C$ and on $D$. Notation-wise, we will specify as an index the round trip on which a cost/benefit is computed. Therefore our goal is, given $\Delta t_C$, $\Delta \hbox{\it Tn}_C$, $\Delta \hbox{\it Pk}_C$ about the round trip on $C$ and likewise $\Delta t_D$, $\Delta \hbox{\it Tn}_D$, $\Delta \hbox{\it Pk}_D$ about the round trip on $D$, to compute $\Delta t_{CD}$, $\Delta \hbox{\it Tn}_{CD}$, and $\Delta \hbox{\it Pk}_{CD}$ about the round trip on $C;D$.
We keep in mind that this computation must be made for any active checkpoint $X$, and therefore in particular for $C$, but also for any other static checkpoint that may occur, any number of times, in $C$, in $D$, or in both.
In addition we will also need to keep track of the basis times and stack sizes when none of the currently activated checkpoints is inhibited i.e. regarding notation, compute $t_{CD}$, $\hbox{\it Tn}_{CD}$, $\hbox{\it Pk}_{CD}$ from $t_C$, $\hbox{\it Tn}_C$, $\hbox{\it Pk}_C$, $t_D$, $\hbox{\it Tn}_D$, and $\hbox{\it Pk}_D$.

%Following Fig.~\ref{figCheckpointModel} again, let us assume that profiling $D$ returns (for each $X$) the cost/benefits $\Delta t$, $\Delta \hbox{\it Tn}$, and $\Delta \hbox{\it Pk}$ for each activated static checkpoint $X$ encountered in $D$, and likewise when profiling $C$. Notation-wise, as cost/benefits can be evaluated independently for each $X$, we will drop the $X$, explicit the $C$ or $D$, obtaining six cost/benefits (for each $X$) to combine: $\Delta t_C$, $\Delta \hbox{\it Tn}_C$, $\Delta \hbox{\it Pk}_C$ and likewise for $D$: $\Delta t_D$, $\Delta \hbox{\it Tn}_D$, $\Delta \hbox{\it Pk}_D$. In addition, profiling will also keep track of the basis times and stack sizes when none of the currently activated checkpoints $X$ is inhibited: $t_D$, $\hbox{\it Tn}_D$, $\hbox{\it Pk}_D$, $t_C$, $\hbox{\it Tn}_C$, and $\hbox{\it Pk}_C$. From these numbers, let us see how profiling will compute $t_{CD}$, $\hbox{\it Tn}_{CD}$, $\hbox{\it Pk}_{CD}$ and (for each $X$) $\Delta t_{CD}$, $\Delta \hbox{\it Tn}_{CD}$, and $\Delta \hbox{\it Pk}_{CD}$.

For the base case of a piece of code that contains no activated checkpoint, profiling can easily return the basis times and stack sizes. Note that $\hbox{\it Pk}$ and $\hbox{\it Tn}$ are then equal. $\Delta$ values are all zero for all $X$.

For the recursive case, it is easy to verify on Fig.~\ref{figCheckpointModel} that, about run time:
\begin{equation} \label{eq41}
t_{CD} = t_1 + t_D + t_2 + t_C
\end{equation}
where $t_1$ is the extra time to take the snapshot and run the primal $C$, and $t_2$ is the time to read the snapshot. Then about stack sizes:
\begin{equation} \label{eq42}
\hbox{\it Tn}_{CD} = \hbox{\it Tn}_D
\end{equation}
\begin{equation} \label{eq43}
\hbox{\it Pk}_{CD} = max(\hbox{\it Pk}_D, \hbox{\it Pk}_C)
\end{equation}
For each $X$, cost/benefit results are the changes in the value of $t_{CD}$, $\hbox{\it Tn}_{CD}$, and $\hbox{\it Pk}_{CD}$, from the reference case (equations \ref{eq41}, \ref{eq42}, \ref{eq43}) with no additional checkpoint inhibited, to the case where only $X$ is switched to inhibited.
These $\Delta$ values are computed differently whether $X$ coincides with $C$ or not. If $X\neq C$ then the effect of inhibiting checkpoint $X$ does not affect $C$ and the resulting computation scheme remains as in Fig.~\ref{figCheckpointModel}. Therefore:
\begin{equation}
\Delta t_{CD} = \Delta t_{D} + \Delta t_{C}
\end{equation}
\begin{equation}
\Delta \hbox{\it Tn}_{CD} = \Delta \hbox{\it Tn}_D
\end{equation}
\begin{equation}
\Delta \hbox{\it Pk}_{CD} = max
\left(
\begin{array}{l}
\!\!\!\hbox{\it Pk}_D + \Delta \hbox{\it Pk}_D\!\!\!\\
\!\!\!\hbox{\it Pk}_C + \Delta \hbox{\it Pk}_C\!\!\!
\end{array}
\right)
- \hbox{\it Pk}_{CD}
\end{equation}
If on the other hand $X = C$, execution when inhibiting $X$ will follow the scheme of Fig.~\ref{figBasisModel}. There is no duplicate execution of the primal $C$ so that time benefit becomes:
\begin{equation}
\Delta t_{CD} = \Delta t_{D} + \Delta t_{C} - t_1 - t_2
\end{equation}
The stack space used by $\overrightarrow{C}$ now accumulates with that used by the round trip $\overline{D}$. Stack size after $\overrightarrow{C}$ in Fig.~\ref{figBasisModel} is indeed $\hbox{\it Tn}_C + \Delta \hbox{\it Tn}_C$. On top of it is accumulated the stack space specifically used by $\overrightarrow{D}$ and $\overline{D}$ which are respectively $\hbox{\it Tn}_D + \Delta \hbox{\it Tn}_D - \hbox{\it Snp}$ and $\hbox{\it Pk}_D + \Delta \hbox{\it Pk}_D - \hbox{\it Snp}$, where {\it Snp} is the stack size measured just after taking the snapshot.
\begin{equation}
\Delta \hbox{\it Tn}_{CD} = \hbox{\it Tn}_C + \Delta \hbox{\it Tn}_C + \Delta \hbox{\it Tn}_D - \hbox{\it Snp}
\end{equation}
\begin{equation} \label{eq49}
\begin{array}{ll}
\Delta \hbox{\it Pk}_{CD} = & \!\!\!max
\left(
\begin{array}{l}
\!\!\!\hbox{\it Tn}_C \!+\! \Delta \hbox{\it Tn}_C \!+\! \hbox{\it Pk}_D \!+\! \Delta \hbox{\it Pk}_D \!-\! \hbox{\it Snp}\!\!\!\\
\!\!\!\hbox{\it Pk}_C + \Delta \hbox{\it Pk}_C
\end{array}
\right)\\
& - \hbox{\it Pk}_{CD}
\end{array}
\end{equation}

All the times and stack sizes needed can be measured while running with $C$ activated.
Note that $\Delta \hbox{\it Tn}_{CD}$ and $\Delta \hbox{\it Pk}_{CD}$ can be negative in the extreme case where $\hbox{\it Tn}_C$ is less than $\hbox{\it SNP}$, or in other words when the snapshot for checkpointing $C$ is larger than the collection of intermediate values stored during $\overrightarrow{C}$. In this extreme case, checkpointing $C$ is never beneficial.

\section{Implementation}\label{secImplementation}

We implemented our profiling tool inside the source-transformation AD tool Tapenade~\cite{TapenadeRef13}.

\subsection{Installing callbacks}

Section~\ref{secProfiling} tells us which elementary data must be collected from the profiled run in order to compute the estimated cost/benefits. These are, for each active checkpoint:
\begin{itemize}
\item the run times $t_1$ and $t_2$.
\item the stack size $Snp$ immediately after taking the snapshot.
\end{itemize}
and in addition every local peak stack size $Tn$, measured at each turn point.
Collecting this data is implemented by installing a number of callbacks in the differentiated code. These callbacks are inserted automatically by Tapenade when invoked with the {\tt -profile} command-line option. Figure~\ref{figProfiledCode} sketches the instrumented adjoint code produced by Tapenade, corresponding to Fig~\ref{figCheckpointModel}. The added callbacks (highlighted in red) all belong to a new {\tt adProfileAdj} package. Calls occur in the forward sweep before writing a snapshot and before and after calling the duplicate primal $C$, and symmetrically in the backward sweep before reading the snapshot and before and after calling the round trip $\overline{C}$ which is $\overrightarrow{C};\overleftarrow{C}$.
Last but not least, a callback to {\tt adProfileAdj\_turn()} is inserted at each turn point between a forward and a backward sweep.
\begin{figure}
\begin{tabular}{|l|}
\hline
\begin{minipage}[t]{8cm}
{\it $<$ code for $\overrightarrow{U}$ $>$}\\
{\color{darkred}\tt CALL adProfileAdj\_SNPWrite("C", 42)}\\
{\tt CALL PUSH({\it snapshot})}\\
{\color{darkred}\tt CALL adProfileAdj\_beginAdvance("C", 42)}\\
{\tt CALL C()}\\
{\color{darkred}\tt CALL adProfileAdj\_endAdvance("C", 42)}\\
{\it $<$ code for $\overrightarrow{D}$ $>$}\\
\end{minipage}\\
\hline
\begin{minipage}[t]{8cm}
{\color{darkred}\tt CALL adProfileAdj\_turn()}\\
\end{minipage}\\
\hline
\begin{minipage}[t]{8cm}
{\it $<$ code for $\overleftarrow{D}$ $>$}\\
{\color{darkred}\tt CALL adProfileAdj\_SNPRead("C", 42)}\\
{\tt CALL POP({\it snapshot})}\\
{\color{darkred}\tt CALL adProfileAdj\_beginReverse("C", 42)}\\
{\tt CALL $\overline{\hbox{C}}$()}\\
{\color{darkred}\tt CALL adProfileAdj\_endReverse("C", 42)}\\
{\it $<$ code for $\overleftarrow{U}$ $>$}\\
\end{minipage}\\
\hline
\end{tabular}
\caption{Adjoint code with profiling callbacks inserted, corresponding to the code sketch of Fig~\ref{figCheckpointModel}. Forward sweep on top, turn point in the middle, backward sweep below. Except for {\tt turn}, callbacks take as argument the called function name and the corresponding line number in the primal code, for later reference.}
\label{figProfiledCode}
\end{figure}

The profiling mechanism requires minor additional adaptations to blend well with sophisticated structures of the adjoint code such as binomial adjoint (for time-stepping loops) and two-phases adjoint (for fixed-point loops).

\subsection{Callback implementation}

Implementation of the profiling callbacks maintains an internal stack of the activated checkpoints encountered, to retrieve correspondence from forward sweep to backward sweep, between the callbacks on a given dynamic checkpoint. Timing $t_1$ is deduced from measurements between {\tt SNPWrite} and {\tt endAdvance}, $t_2$ between {\tt SNPRead} and {\tt beginReverse}. The final computation of the cost/benefits for the round trip $C;D$ is entirely performed during {\tt endReverse}, when all the needed values $t$, $\hbox{\it Tn}$, $\hbox{\it Pk}$, and all $\Delta t$, $\Delta \hbox{\it Tn}$, $\Delta \hbox{\it Pk}$ are known for $C$ and $D$.
At the end of the profiled run, a readable summary of the cost/benefits is displayed.

Complexity of the profiling process lies in that of callback {\tt endReverse}, that implements all equations \ref{eq41} to \ref{eq49}, and that is called once per run-time procedure call (or encounter of a manual checkpoint). In each {\tt endReverse}, equations \ref{eq41}, \ref{eq42}, and \ref{eq43} need to be evaluated only once, whereas the subsequent equations need to be evaluated once per static checkpoint $X$. The total complexity is therefore the product of the size of the call tree times the number of static checkpoints. Memory-wise, the data stored by profiling occupies a size at most proportional to the number of static checkpoints times the depth of the call tree.

\subsection{Inhibiting / activating checkpoints}

Bearing in mind that calls are always checkpointed by default, the functionality for inhibition of potential checkpointing locations was already present in the AD tool Tapenade. End-users could play with it but our impression is that very few did so, possibly due to the absence of a profiling tool to guide them. The user interface of Tapenade provides two ways to inhibit checkpointing on calls:
\begin{itemize}
    \item On the Tapenade command-line, one can use option {\tt -nocheckpoint "foo foo2"}, which will prevent checkpointing on every call to procedures {\tt foo} and {\tt foo2}.
    \item In the primal source, one can place pragma {\tt \$AD NOCHECKPOINT} before definition of a procedure {\tt foo}, which inhibits checkpointing on every call to {\tt foo} (equivalent to the command line option). Alternatively one can place the same pragma before any individual call to {\tt foo}, thus selectively inhibiting this particular call.
\end{itemize}

\subsection{Inhibiting checkpointing on externals}
Since external routines are not shown to the AD tool, Tapenade provides a way for the end-user to specify information about these externals that are useful for a better differentiation, through a special-purpose specification file. These are typically about types and read/write status of their arguments, or about the differentiable dependency of their outputs on their inputs. The end-user is also required to write the derivative code for these externals. For a non-checkpointed call to an external, the AD tool can produce a better code by knowing which of the arguments are used in a ``{\em non-linear}'' way by the external, or in other words which of the values {\em returned} by the external must be, when later overwritten, preserved on the stack for future use in the backward sweep. If this information is not provided to the AD tool, the conservative information is built from the read/write status, possibly causing stack storage of more values than necessary. We thus extended the mini-language of the specification file to designate these ``non-linear'' arguments.

Symmetrically, the differentiated code for a non-checkpointed external {\tt foo} consists of two procedures {\tt foo\_fwd} and {\tt foo\_bwd} (instead of a single {\tt foo\_b} in the classical checkpointed case). The end-user is required to provide their implementation. One consequence of not checkpointing is that {\tt foo\_fwd} may be required by the differentiation context to take care of preserving some of its input values, on the stack, and {\tt foo\_bwd} must restore these values from the stack at the symmetric location. This extra task for the end-user is easy enough and worth the effort. However, in cases where the user can not do so, the conservative fallback option is to let the AD tool store what is needed immediately before the call to {\tt foo\_fwd} and restore just after {\tt foo\_bwd}. This will be triggered by another option in the external specification file. Interaction with the end-user thus occurs in two phases:
\begin{enumerate}
    \item Before differentiation, the end-user can use the specification file to provide the non-linearly used arguments of the external procedures. The user also has the option to accept that the hand-written {\tt foo\_fwd} and {\tt foo\_bwd} take care of storing and restoring some of their arguments.
    \item After differentiation, the AD tool emits messages that tell the user which differentiated external procedures must be provided, specifying which derivatives (of which output with respect to which input) must be computed, and whether they must come in the {\tt foo\_b} form or in the {\tt foo\_fwd}, {\tt foo\_bwd} form. When the user has accepted to do so, messages also specify the input arguments that must be preserved in case they are overwritten in {\tt foo\_fwd}.
\end{enumerate}

\subsection{Asynchronous stack offloading to files}

Searching for a good checkpointing scheme is all the more important for large applications that need a huge amount of memory to store the stack. Until recently Tapenade kept this stack in RAM, and this has been noted as an unfortunate limitation by several users. We implemented a new stack mechanism in Tapenade, that triggers offloading of the deeper parts of the stack to files. It is a variation on virtual memory: when the stack size grows close to a user-defined given limit, the parts of the stack that were pushed early, and therefore that will be popped late, are stored to files and the associated main memory becomes available again. The reciprocal mechanism occurs during the backward sweep. These deep parts of the stack, that are subject to this file write/read mechanism, are in general ``dormant'' i.e. disjoint from the stack top where all push/pop activity occurs. Therefore Tapenade offers the option of performing the file write/read operations asynchronously, using Posix {\tt pthreads}. When a spare thread is available, this reduces the overhead of the file offloading mechanism.

For instance the application described in section~\ref{secbiogeo} makes use of this stack offloading mechanism, with a stack peak that goes over 4 Gb. while never using more main memory than a user-defined limit of 256 Mb.

\section{Application}\label{secApplication}

\ssg{We test our profiling capability on test cases from the Massachusetts Institute of Technology General Circulation Model (MITgcm) \cite{Marshall97,mitgcm_doc}. The MITgcm is widely used by the climate science community to simulate planetary atmosphere and ocean circulations, and its dynamical core has been successfully used to also perform ice stream and ice shelf simulations, and thus develop synchronous ice-ocean coupling capabilities as well as its adjoint \cite{Jordan2018, Goldberg2018}.}

\ssg{MITgcm has always been developed to be compatible with AD tools, specifically TAF~\cite{Giering1998}, enabling the generation of tangent-linear and adjoint models. It is thus a key component of the Estimating the Circulation and Climate of the Ocean (ECCO) state estimate: a data assimilation product widely employed by the physical oceanography research community. The MITgcm has recently been differentiated using Tapenade \cite{Gaikwad2024}, which is what allows us to perform the profiling experiments discussed next.}

\subsection{Test {\tt streamice}:}
\label{secStreamice}
\ssg{The \texttt{halfpipe\_streamice} test case uses the package \texttt{streamice} to simulate the flow of land ice in a valley. The simulation runs for 80 timesteps on a  $40 \times 20$ horizontal grid. The individual time steps are more complex than a typical MITgcm simulation since they contain fixed point iteration based solvers to solve the non-linear stress balance. Their adjoint is built according to Christianson's two-phases algorithm \cite{Christianson1994,Goldberg2016}.}

Adjoint AD of {\tt streamice} was first performed with OpenAD in 2016~\cite{Goldberg2016}, and a few years later with Tapenade. Both tools apply the same strategy for the adjoint of the fixed-point loop. In both cases the end-user chose not to apply binomial checkpointing to the outer time-stepping loop. The structure of the adjoint codes is therefore similar. Typically, run-time performance of an adjoint code is measured by the ratio with the run-time of the primal. The adjoint code by Tapenade yielded a 7.0 ratio, which is fair but not as good as with OpenAD.
%{\color{red}[From JLB:] Can we add a ref to these figures? (for instance what was presented at ECCO just over a month ago?) SSG: Maybe reference is not needed since a user can just run the experiment using a single testreport command.}
Likewise, on most MITgcm test cases, the ratio obtained with Tapenade is not as good as the average ratio obtained with TAF. By default, Tapenade activates checkpointing at each procedure call, whereas OpenAD doesn't. Checkpointing in TAF is carefully hand-tuned, with methods that have not been published. Certainly, there is room for improvement from the default checkpointing scheme of Tapenade.

\begin{figure}
\includegraphics[width=\linewidth]{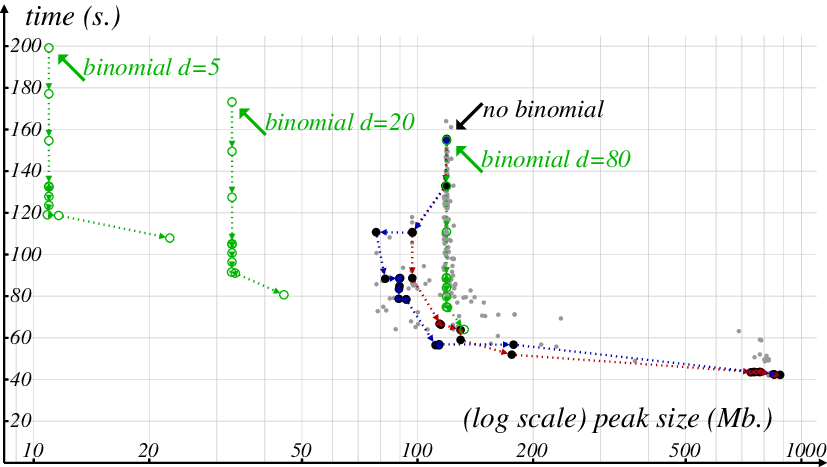}
\caption{Tuning the adjoint of {\tt streamice} by profiling. Each checkpointing configuration is shown as a dot whose coordinates are peak stack size and run time of the adjoint. The red and blue dotted lines show the evolution of performance of the adjoint by repeatedly following the suggestions made by our profiling tool. The red line follows a ``run-time gain first'' strategy, the blue line a ``careful on memory'' strategy. Time measurements are averaged on 5 runs. Smaller gray dots show the performance of 250 random checkpointing configurations. Green dots show combination with binomial checkpointing, discussed in section~\ref{secWithBinomial}}
\label{figExpeStreamice}
\end{figure}

Fig.~\ref{figExpeStreamice} summarizes this search for a better checkpointing configuration. The initial configuration with all checkpoints activated (indicated as ``{\it no binomial}'') has an average run-time of 155.4 seconds and a peak stack size of 119.1 megabytes. The number of static checkpoint locations is 85, yielding an enormous $2^{85}$ number of possible configurations to explore.

We use our profiling tool to explore only the most profitable changes. Incidentally, we compare adjoint run-time with and without profiling and find a negligible 1.1 second overhead for profiling. The suggestions returned by the profiled adjoint code are sorted in 3 categories:
\begin{itemize}
    \item Static checkpoints whose inhibition give a run-time gain (as always), and a stack peak gain. Those are rare cases where the snapshot is larger than the amount of intermediate values stored in the forward sweep. This is generally due to overestimation of the AD data-flow analysis on arrays, where activity for a single cell of an array implies activity of the whole array. This sort of profiling result means that the checkpoint should be inhibited without hesitation. 
    \item Static checkpoints whose inhibition give no change of the stack peak. This happens when the effect of inhibiting the checkpoint indeed modifies the size of one local peak of the stack, but this peak is smaller than the global peak and therefore the change is hidden. The checkpoint is a good candidate for inhibition, as this will gain some run time without changing the global peak size.
    \item Static checkpoints whose inhibition will increase the stack peak size. Decision to inhibit it results from a trade off decision, comparing with the run-time expected benefit and possibly the available budget in memory space.
\end{itemize}
As these suggestions allow for several alternative decisions, the end-user may apply different strategies. Two example strategies are presented in Fig.~\ref{figExpeStreamice}: one (red line) follows the suggestions with the highest run-time gain first, but deferring as much as possible suggestions with a high memory cost. The other (blue line) follows suggestions with the lowest memory cost first, regardless of run-time gain. For both strategies, only one or a few suggestions are selected and applied jointly, then differentiation is performed and the adjoint code is run again, yielding new suggestions. This process is repeated until no suggestion remains, which is when all 85 checkpoints are inhibited. In real life, the end user will probably stop the process before that stage, as the memory cost may go over the available memory budget. Fig.~\ref{figExpeStreamice} shows with black dots the time and memory performance of the successive adjoint codes after each step. Gray dots show the performance of 250 randomly chosen checkpointing configurations. We observe that they are mostly situated above and right of the red and blue lines, indicating that the profiling suggestions, combined with these strategies, take performance close to the front of optimal configurations.

From the results in Fig.~\ref{figExpeStreamice}, we find a few checkpointing configurations that offer a significant time gain for a relatively small memory cost, or even sometimes a small gain. For instance one dot on the red line exhibits a stack peak of 176.0 Mb, not too much above the initial 119.1 Mb, and a time of 51.9 seconds, a significant gain from the initial 155.4 s. resulting in a slowdown ratio of 2.4. If memory is really scarce, another dot on the blue line exhibits a stack peak of 111.41 Mb, even less than the initial 119.1 Mb, and a time of 56.49 s. and therefore a still very good ratio of 2.6. Unless memory is very abundant, it is unlikely that the end-user pushes the process to its extreme, costing 880.6 Mb, even if time becomes as short as 42.2 s.

\subsection{Binomial checkpointing}
\label{secWithBinomial}

A natural question is the comparison and interaction between our checkpointing on the call tree and binomial checkpointing on time-stepping loops.
Let us introduce binomial checkpointing into the {\tt streamice} test case by adding directive
\centerline{\tt \$AD BINOMIAL-CKP}
on the time-stepping loop. Although the test case features a limited number of time steps (80), our experiments clearly capture the effect of binomial checkpointing and its interaction with our profiling method. Binomial checkpointing is tuned by choosing the maximum number $d$ of ``before time-step'' snapshots that can be stored together in memory. Figure~\ref{figExpeStreamice} shows (in green) experiments with an extreme $d=80$, a more reasonable $d=20$, and a quite realistic $d=5$. For each $d$, dotted arrows  show the evolution of performance when following profiling suggestions (highest run-time gain first) until no suggestion remains. The $d=80$ case is interesting mostly as it makes the link with the experiment with no binomial ckeckpointing: allowing $d$ to be as high as the total number of time-steps is equivalent to discarding binomial checkpointing altogether and activating checkpointing on the time-stepping loop body, which happens here to be a single procedure call. Therefore all green dots of the $d=80$ evolution are equivalently checkpointing configurations of the experiment with {\em no} binomial checkpointing (i.e. black and small gray dots), but they belong to a subset where checkpointing remains activated on the time-stepping loop body. Evolution of the green dots is therefore restricted and cannot reach performance as good as the red and blue evolutions. Notice that the red and blue evolutions almost immediately choose to inhibit checkpointing on the time-stepping loop body.

Binomial checkpointing controls memory usage through the choice of $d$. A smaller $d$ reduces the memory used to store snapshots, at the cost of a higher number of duplicate execution of time-steps. The green dots for $d=20$ and $d=5$ demonstrate that. For all three $d$, our profiling tool provides profitable suggestions, decreasing run-times by a factor 1.8 to 2.4 at almost no memory cost.

We interpret the almost vertical shape of the green curves by the fact that each configuration systematically uses $d$ ``time-stepping'' snaphots for any activation pattern of the other checkpoints.
Observe that with binomial checkpointing, round trips on each time-step always occur separately.
Therefore, the memory cost of inhibiting checkpoints inside a time-step is not multiplied by the number of time steps.
Profiling suggestions reflect that, by finding lower memory costs (and benefits) to inhibiting checkpoints that are contained in a time-stepping loop with binomial checkpointing. 
As long as this cost is small compared to that of the $d$ ``time-stepping'' snaphots, it remains almost invisible.
It only becomes visible when almost all checkpointing locations are inhibited. 

\subsection{Test {\tt tutorial\_global\_oce\_biogeo} :}
\label{secbiogeo}

\begin{figure}
\includegraphics[width=\linewidth]{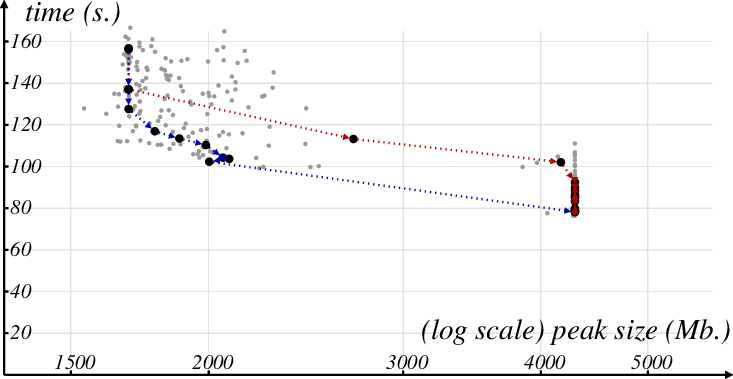}
\caption{Tuning the adjoint of {\tt tutorial\_global\_oce\_biogeo} by profiling. Each checkpointing configuration is shown as a dot whose coordinates are peak stack size and run time of the adjoint. The red and blue lines show the evolution of performance guided by profiling, respectively with a ``run-time gain first'' strategy, and a ``careful on memory'' strategy. Gray dots show the performance of 250 random checkpointing configurations.}
\label{figExpeBiogeo}
\end{figure}

\ssg{The MITgcm test case {\tt tutorial\_global\_oce\_biogeo} is a global simulation on a $128\times64\times15$ grid. It also simulates the biogeochemistry aspects of the ocean, for example, the carbon uptake and how it is affected by the sea surface temperature anomalies. We simulate $10$ steps so that the runtime is similar to the {\tt halfpipe\_streamice} test. There is no binomial checkpointing as a default.} Figure~\ref{figExpeBiogeo} shows the improvement in performance obtained by applying the suggestions from profiling, with the same two strategies as in Fig~\ref{figExpeStreamice}.

From this experiment it becomes clear that making an informed choice of checkpointing is critical. Indeed, one can easily reduce by 35\% the compute time, yet avoiding the massive memory consumption of an extreme ``no checkpoint'' strategy.

\section{Further research}\label{secFurther}

Experience raises a few questions about our approach. One issue is that profiling is run on one given checkpointing configuration, and changing this configuration can lead to another regime with different profiling results, although often only slightly. Therefore it may be safer to re-run differentiation and profiling each time one checkpoint is inhibited. This can prove impractical and we often chose to switch several checkpoints at a time, possibly missing a better configuration. In other words, to which extent can we use one set of profiling suggestions for several checkpoints at a time, or re-use this set even if some more checkpoints are inhibited? This may relate to relative position in the tree of nested round trips. 

Alternatively, one can probably extract enough information from the initial profiling run, enabling some separate optimal search tool to directly find the front of ``optimal'' configurations. This could avoid the need for re-profiling, but would require storage of profiling data proportional to the size of the (dynamic) tree of nested checkpoints, which can be large. In comparison, our approach only stores at a given time data proportional to the depth of the tree of nested checkpoints, times the number of static checkpoints.

Profiling predictions can also be inaccurate due to optimization of the adjoint code, which the AD tool performs using static data-flow analysis. This leads in many cases to inexact predictions, often too optimistic in favor of inhibiting a checkpoint. To understand the issue, recall that an activated checkpoint $C$ causes the forward sweep $\overrightarrow{C}$ to be followed immediately by $\overleftarrow{C}$. A good AD tool will take advantage of this to remove some ``{\em adjoint-dead}'' code at the junction of the two sweeps, such as intermediate values pushed then immediately popped from the stack, as well as primal computations that are not needed for the upcoming backward sweep. Taking the snapshot before running the primal $C$ also has the consequence of possibly reducing the set of values that will be stored during $\overrightarrow{D}$ and $\overrightarrow{C}$ (found by the ``TBR'' analysis) because the snapshot already stored these values. The benefits of these data-flow analysis is invisible at profiling time. All in all, time and memory performance of the configuration with an activated checkpoint $C$ are not easily modeled, or equivalently inhibiting $C$ often brings improvements that are not exactly those deduced from our profiling measurements. An obvious answer to this disturbing issue would be to de-activate these data-flow analysis in the AD tool, at least during the iterative profiling process. However, one would then optimize a derivative code which is not the same as the final one, with a chosen checkpointing configuration that may not be the most appropriate.

To validate the discussion above, we have run two series of profiling experiments on the {\tt streamice} test case, that are identical except that one series ({\it optim-on}) is performed with the standard Tapenade whereas the other ({\it optim-off}) uses a Tapenade with a deactivated ``adjoint-dead'' optimization. We could not deactivate the ``TBR'' analysis, as this changes the adjoint code  too radically. In each series of experiments, we have considered the following cost/benefit predictions: on the one hand each prediction corresponding to each suggestion followed along the ``careful on memory'' itinerary (blue line of Fig.~\ref{figExpeStreamice}), and on the other hand all predictions, followed or not, made by the initial profiling step i.e. on the initial configuration with all potential checkpoints activated. For technical reasons, predictions made about procedures called at several static locations are not considered. For each prediction, we inhibit the corresponding checkpoint, then differentiate and run, thus measuring the actual peak stack change for comparison with the predicted one.
Our results shown in Table~\ref{table:nooptim} mostly corroborate our tentative explanation as we observe better predictions on the {\it optim-off} experiment. Still, this deserves more investigation.

\begin{table}
\centering
\begin{tabular}{l||r|r|r|r}
\hline
& exact & good & acceptable & inexact \\
\hline
\hline
{\it optim-on}  & 24 & 9 & 17 & 28 \\
{\it optim-off} & 29 & 17 & 11 & 21 \\
\hline
\end{tabular}
\caption{Exactness of memory cost predictions when activating or deactivating the ``{\it adjoint-dead}'' adjoint code optimization. In both settings, we count the number of predictions that are ``exact'', not exact but ``good'' (estimation error less than 0.01Mb), not good but ``acceptable'' (estimation error less than 0.1Mb).}
\label{table:nooptim}
\end{table}

It is visible in figures~\ref{figExpeStreamice} and~\ref{figExpeBiogeo} that some relatively better checkpointing configurations are not attained by following the two strategies that we experimented with. Visually there are a few gray dots below-left of the dotted lines. We mostly interpret this as a consequence of the few inexact profiling predictions that we just mentioned. It also indicates that, in addition to the red and blue dotted lines on the figures, there may exist better strategies for selecting the next profiling suggestion to follow.

As already mentioned, checkpointing can also be applied at the level of time-stepping loops, for which the well-known binomial checkpointing~\cite{Griewank2000ARA} provides an optimal schedule. On large codes, we need to apply both ``kinds'' of checkpointing: (optimal) binomial checkpointing on time-stepping loops, and call tree checkpointing (tuned by profiling) on the rest of the code. Unfortunately we have no unifying framework that encompasses the two. Nevertheless, interactions exist between both: the same stack that we used throughout this work can also accommodate the snapshots of binomial checkpointing. Given a memory budget for this stack, the choice exists to use it to allow for more snapshots for binomial checkpointing, or to inhibit more call-tree checkpoints in the body of the time-steps. Since a cost model exists for both options, the profiling tool could suggest which option is more profitable, thus making one step towards blending the two approaches.

\section{Conclusion}\label{secConcl}

The memory/recomputation trade off known as checkpointing is a key ingredient for adjoint (or reverse) differentiation of large applications. Checkpointing can and must be applied at many levels in the target application, in particular at the nodes of the call tree or on iterative loops. The choice of the locations that deserve checkpointing or not has a major impact on the final performance, in run time and memory usage, of the adjoint code. Yet very little guidance is provided to the end-user. We proposed a profiling algorithm that predicts the performance cost/benefit of checkpointing for each node of the call tree. We experimented with greedy strategies that exploit these predictions, progressively improving performance of the adjoint code by adapting its checkpointing schedule. On the two large test cases that we studied, both from the MITgcm test suite, we obtain significant performance gains, up to a two- or three-fold speedup for a minor memory cost. Profiling for better checkpointing appears as very promising.

The profiling algorithm that we proposed is a prototype. It is now available as a component of the AD tool Tapenade. It could be worth exploring more sophisticated strategies to find almost optimal checkpointing schedules, directly instead of step by step, from a similar profiling pass. It also seems profitable to guide the end-user decision better based on profiling results, or to further automate the process. In particular the interplay between call tree checkpointing and binomial checkpointing should be quantified as well, and integrated in the decision strategy.

\section*{Acknowledgments}

\bibliographystyle{plain}
\bibliography{profiling}

\end{document}